%% file: main.tex
\documentclass[a4paper, 10pt, conference]{ieeeconf}      
\usepackage{FG2021}

\FGfinalcopy 

\IEEEoverridecommandlockouts                              
\usepackage{graphics} 
\usepackage{epsfig} 
\usepackage{amsmath} 
\usepackage{amssymb}  
\usepackage{booktabs} 
\usepackage{mathrsfs}
\usepackage[ruled, linesnumbered]{algorithm2e}
\usepackage{algorithmic}
\usepackage{hhline}
\usepackage{multirow}
\usepackage{array}
\usepackage{balance}
\usepackage{float}
\usepackage{hyperref}
\newcolumntype{P}[1]{>{\centering\arraybackslash}p{#1}}
\usepackage{afterpage}

\usepackage[accsupp]{axessibility}

\def\FGPaperID{275} 

\title{\LARGE \bf
Spatially Constrained GAN for Face and Fashion Synthesis
}

\author{\parbox{16cm}{\centering
    {\large Songyao Jiang$^1$, Hongfu Liu$^2$, Yue Wu$^1$ and Yun Fu$^{1,3}$}\\
    {\normalsize{
    $^1$Department of Electrical and Computer Engineering, Northeastern University, Boston MA, USA\\
    $^2$Michtom School of Computer Science, Brandeis University, Waltham MA, USA\\
    $^3$Khoury College of Computer Science, Northeastern University, Boston MA, USA}}}
}

\newcommand{\ie}{\textit{i}.\textit{e}., }
\newcommand{\eg}{\textit{e}.\textit{g}., }

\begin{document}

\ifFGfinal
\thispagestyle{empty}
\pagestyle{empty}
\else
\author{Anonymous FG2021 submission\\ Paper ID \FGPaperID \\}
\pagestyle{plain}
\fi
\maketitle

\begin{abstract}
Image synthesis has raised tremendous attention in both academic and industrial areas, especially for conditional and target-oriented image synthesis, such as criminal portrait and fashion design. The current studies have achieved encouraging results along this direction, but they mostly focus on class labels where spatial contents are randomly generated from latent vectors. Some recent studies have explored spatial constraints for generative models guided by semantic segmentation, but most of them are designed for scene generation and lack random variation. Such methods are not suitable for face or fashion image synthesis, where different images may share the same semantics. Different from all the current methods, we decouple the image synthesis task into three independent dimensions and propose a novel Spatially Constrained Generative Adversarial Network (SCGAN) to model it. SCGAN uses a simple yet effective way to decouple spatial constraints and attribute conditions from latent vectors, and treat them as additional controllable signals via a segmentor and a specially designed generator. Other unregulated contents are left to be generated from latent vectors. Experimentally, we provide both qualitative and quantitative results on CelebA and DeepFashion datasets to demonstrate that the proposed SCGAN is very effective in synthesizing spatially controllable and attribute-specific images with high visual quality and large variations.
Our code is provided at 
\url{https://github.com/jackyjsy/SCGAN}.
\end{abstract}

\input{SCGAN_body.tex}





{\small\balance{
\bibliographystyle{ieee}
\bibliography{SCGAN_short}
}}

\end{document}

%% file: SCGAN_body.tex
\section{Introduction}
The success of Generative Adversarial Networks (GAN)~\cite{goodfellow2014generative} upsurges an increasing trend of realistic image synthesis~\cite{zhao2017stylized,zhang2017stackgan,wang2018high}, where a generator network produces artificial samples to mimic the real samples from a given dataset and a discriminator network attempts to distinguish between the real samples and artificial samples. These two networks are trained adversarially as two players in a game, and eventually, the two-player game will end up with the Nash Equilibrium. In such equilibrium, the generator is capable of mapping latent vectors from a simple distribution to real data samples from a complex distribution, while the discriminator can hardly distinguish the artificial samples from the real ones. GANs have been widely used in many applications such as natural language processing~\cite{zhang2016generating,yu2017seqgan}, image super-resolution~\cite{ledig2016photo,liu2017beyond}, domain adaptation~\cite{hoffman2016fcns,bousmalis2017unsupervised}, object detection~\cite{li2017perceptual}, activity recognition~\cite{li2017region}, video prediction~\cite{mathieu2015deep}, face aging \cite{liu2017face}, semantic segmentation~\cite{luc2016semantic}, face frontalization~\cite{yin2020dual,yin2021superfront}, and image translations~\cite{isola2016image,zhu2017unpaired,jiang2021geometrically}. 

\begin{figure}[t]
  \centering
  \includegraphics[width=0.48\textwidth]{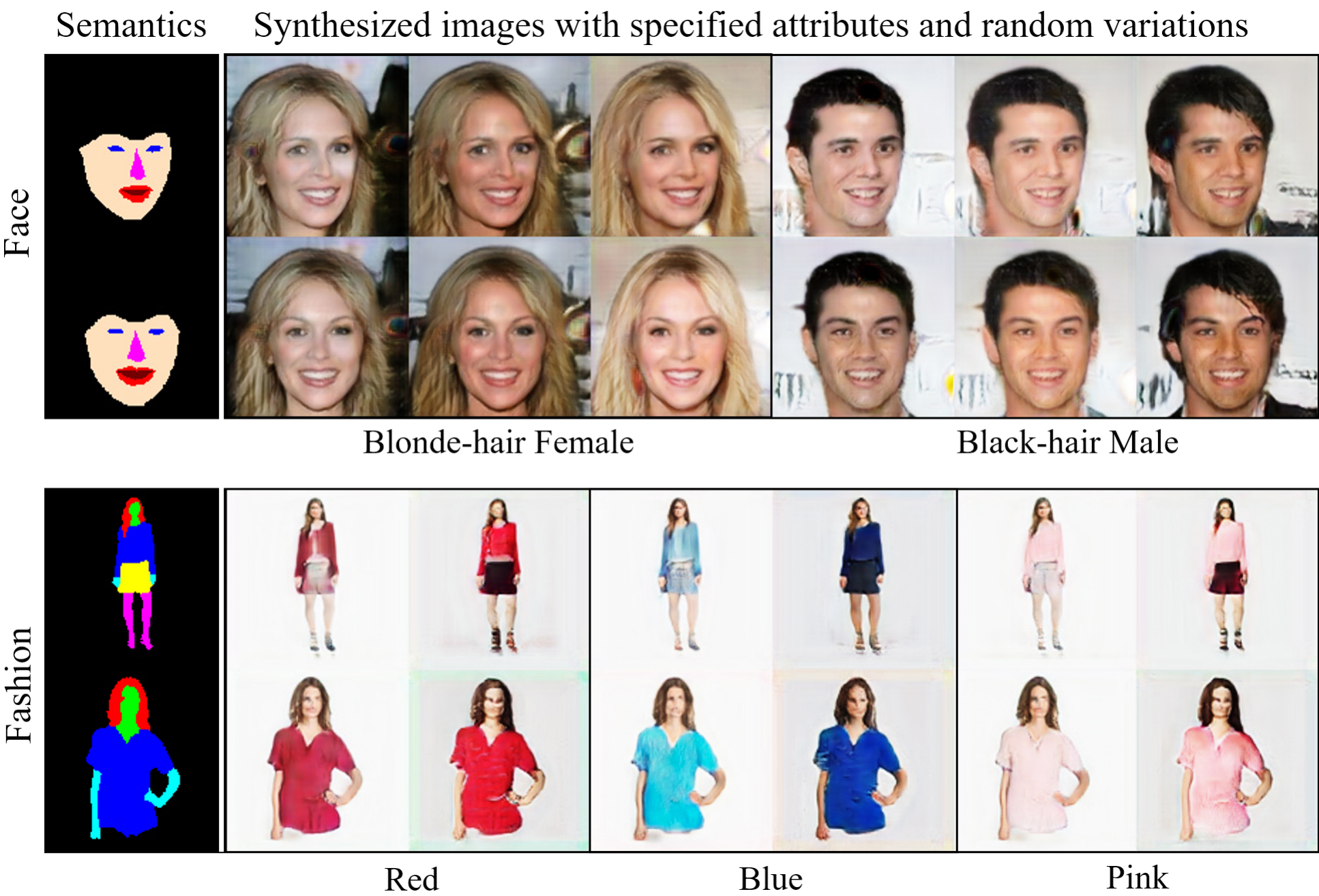}\vspace{-0.15cm}
  \caption{SCGAN decouples the image synthesis task into three dimensions (\ie spatial, attribute and latent dimensions). SCGAN synthesizes face and fashion images guided by target semantic segmentations, specified attributes and achieves large variations on other unregulated components (\eg textures, skin colors, hair styles, fashion design, and color shades).}\label{fig_title}\vspace{-0.3cm}
\end{figure}

Beyond generating arbitrary images, conditional and target-oriented image generation is highly needed in various practical scenarios, such as criminal portraits based on victims' descriptions, clothing design with certain fashion elements, data augmentation, and artificial intelligence imagination. 
cGAN~\cite{mirza2014conditional} first provided a way of conditional generation according to input class labels, which is further extended by \cite{odena2016conditional} and \cite{li2017triple} that additional classifiers are utilized to guide the image generation. They focus on available class labels as the condition where spatial contents are still randomly constructed from latent vectors. The edge details are usually blurred and the boundary information is difficult to preserve due to the lack of spatial constraints. Semantic-guided image synthesis has been recently explored in~\cite{park2019semantic,tang2020local} for scene image synthesis. Those methods use image-to-image translation networks to generate scene images from semantic segmentations. However, when applied to face and fashion image synthesis, those methods cannot provide much diversity with given semantic segmentations. In other words, they are deterministic and tend to synthesize fixed outputs with given input semantics. Some efforts such as SPADE~\cite{park2019semantic} encode a style image to a style vector to obtain diverse outputs. Such design works well for scene images, however, our experiments reveal that such method does not provide a good diversity for face or fashion synthesis. 

For face and fashion synthesis, inherently, there exists a one-to-many mapping from semantic segmentations to real images. Many distinct faces and clothes could share very similar semantics but retain diverse textures and attributes. This is a major reason why those image-to-image translation-based semantic-guided image synthesis methods are not suitable for face and fashion synthesis tasks. To solve the problem, we propose to decouple the face and fashion synthesis tasks into three dimensions, which are spatial dimension, attribute dimension, and latent dimension, and make the first two dimensions explicitly controllable. The spatial configurations of generated images are regulated by input semantic segmentations, the attributes are specified by input attribute labels and the other uncontrolled components are automatically synthesized from input latent vectors. 

We propose a Spatially Constrained Generative Adversarial Network (SCGAN) to learn the mapping of the three-dimension image synthesis. SCGAN consists of three networks, a generator network, a discriminator network with an auxiliary classifier, and a segmentor network, which are trained together adversarially. The generator is specially designed to take a semantic segmentation, a latent vector, and an attribute label as inputs step by step to synthesize a fake image. The discriminator network tries to distinguish between real images and generated images as well as classifying them into multi-label attributes. The discrimination and classification results guide the generator to synthesize realistic images with correct target attributes. The segmentor network attempts to estimate semantic segmentations on both real images and fake images to deliver estimated segmentations, which guides the generator in synthesizing spatially constrained images. With those networks, the proposed SCGAN generates realistic and diverse face and fashion images guided by input semantic segmentations and attribute labels, which enables many interesting applications such as interpolating between left faces and their faces, and generating intermediate faces from not smiling to smiling facial expression. Experimentally, we demonstrate the effectiveness and benefits of the spatial constraints by providing both qualitative and quantitative results on a face dataset CelebA \cite{liu2015faceattributes} and a fashion dataset DeepFashion~\cite{liu2016deepfashion}. Here we highlight our major contributions as follows.

\begin{itemize}
\item We decouple the face and fashion synthesis task into three dimensions (\ie spatial, attribute, and latent) and propose a novel SCGAN to model it. Both spatial and attribute dimensions can be explicitly controllable. 
\item A generator network is particularly designed to extract spatial information from input segmentation, then concatenate a latent vector to provide variations and apply specified attributes. A segmentor network is introduced to guide the generator with spatial information and increases the model stability for convergence.
\item Extensive experiments are conducted on the CelebA and DeepFashion datasets to demonstrate that the proposed SCGAN is effective in controlling spatial and attribute contents and can synthesize face and fashion images with large variations.
\end{itemize}

\begin{figure*}[t!]
  \centering
  \includegraphics[width=0.9\textwidth]{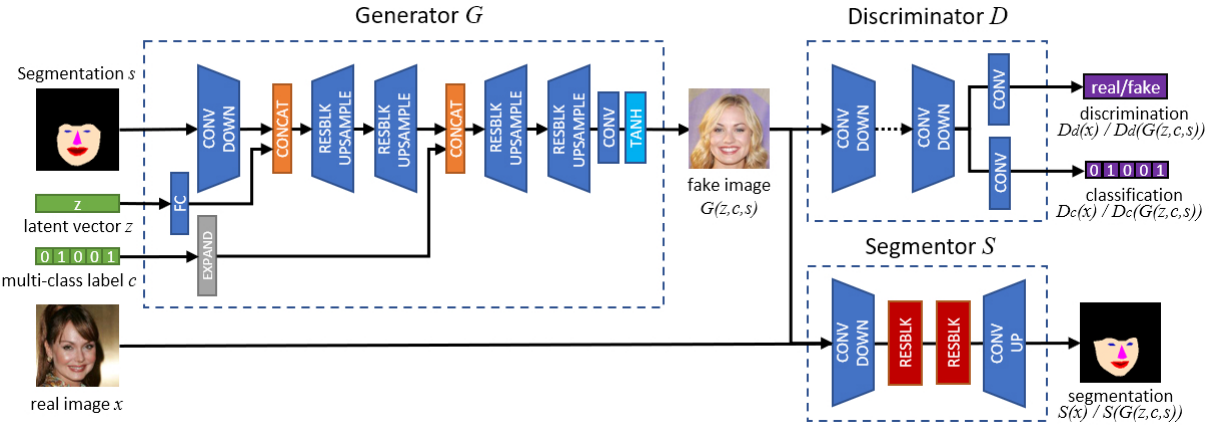}
  \caption{SCGAN consists of a generator, a discriminator with an auxiliary classifier and a segmentor which are trained together. The generator is particularly designed that a semantic segmentation, a latent vector and an attribute label are input to the generator step by step to generate a fake image. The discriminator takes either a fake or real image as input and outputs a discrimination result and a classification result. The segmentor takes either fake or real image as input and outputs a segmentation result, and guides the generator to synthesize fake images which comply with the target segmentation.}\label{fig_framework}\vspace{-0.3cm}
\end{figure*}

\section{Related Work}
\label{section_related}
In recent years, deep generative models inspired by GAN enable computers to synthesize new samples based on the knowledge learned from given datasets. There have been many variations of GAN to improve the generating ability and stabilize adversarial training such as~\cite{radford2015unsupervised,berthelot2017began,arjovsky2017wasserstein,gulrajani2017improved,liu2016coupled,miyato2018spectral,bora2018ambientgan,karras2018progressive,karras2019style,karras2021alias}. 
In the meanwhile, many researchers focused on developing target-oriented generative models instead of random generation. Conditional GAN~\cite{mirza2014conditional} is the first attempt to input conditional labels into both generator and discriminator to achieve conditional image generation. Similarly, ACGAN~\cite{odena2016conditional} constructs an auxiliary classifier within the discriminator to output classification results and TripleGAN~\cite{li2017triple} introduces a classifier network as an extra player to the original two players setting. But all these studies focus on attribute-level conditions and neglect spatial conditions, which leads to the lack of spatial controllability in synthesized images. 

People have been working on manipulating spatial contents of images via 3D morphable models since 1990s~\cite{blanz1999morphable}. Recently, synthesizing spatially constrained images via a GAN-based network is first exploited using image-to-image translation methods, where input images can be regarded as spatial conditions in image translation. Pix2Pix~\cite{isola2016image} is the first to use an image as the conditional input and trains their networks with supervision from paired images. Then many researchers find out that paired training is unnecessary after introducing a cycle-consistency loss and propose several unpaired image translation methods \cite{zhu2017unpaired,kim2017learning,yi2017dualgan,liu2017unsupervised,wang2018high,jiang2019segmentation}. Based on those two-domain translation methods, StarGAN~\cite{choi2017stargan} proposes a multi-domain image translation network with an auxiliary classifier. Vid2Vid~\cite{wang2018video} further extends the image translation to a video translation, which enables many interesting applications such as synthesizing dance videos from skeleton videos. Human body pose landmarks are used as spatial constraints to guide generative networks and synthesize whole-body images in~\cite{hamada2018full, dong2018soft}. 

Most recently, MaskGAN~\cite{lee2020maskgan} utilizes facial attribute masks to enable interactive face image manipulation, SPADE~\cite{park2019semantic} proposes a spatially-adaptive normalization to effectively generate high-resolution images based on given semantic segmentation with different learned styles, and LGGAN~\cite{tang2020local} further improves the ability of semantic-guided scene generation to synthesize small objects and detailed local textures. Style-guided image translation methods~\cite{choi2020stargan,zhu2020sean} merge a base image with a style image to synthesize a new image that spatial configuration and attribute-level contents are inherited and not explicitly controllable. All the above methods have intrinsic assumptions of one-to-one mappings that they synthesize deterministic images without much variation. Therefore, those methods perform quite well for scene synthesis but are not suitable for conditional face or fashion generation which require large variations. 

Different from all existing methods, SCGAN decouples the image generation into three dimensions via a simple and novel design of networks, and utilizes semantic segmentation as spatial constraints in a distinctive way. SCGAN takes a latent vector, attribute labels, and a semantic segmentation as inputs, explicitly controls spatial configurations and attribute contents, and generates target images with a large diversity.

\section{Methodology}
\label{section_method}
In this section, we first define our target problem and define the symbols used in our methodology. Then we introduce the framework structure of the proposed SCGAN. After that, all loss terms in the objective functions to optimize those networks are discussed in detail. Last, we provide a detailed training algorithm.

\subsection{Problem Setting}
Let $\mathbb{P}(x,c,s)$ denotes the joint distribution of the target joint dataset with attribute labels and geometric configuration, where $x$ is a real image of size ($H \times W \times 3$), $c$ is its multi-attribute label of size ($1 \times n_c$) with $n_c$ as the number of attributes, and $s$ is its semantic segmentation of size ($H \times W \times n_s$) with $n_s$ as the number of segmentation classes. Each pixel in $s$ is represented by an one-hot vector with dimension $n_s$, which codes the semantic index of that pixel. Our goal can be described as finding the mapping $G\left(z,c,s\right)\rightarrow y$, where $G(\cdot,\cdot,\cdot)$ is the generating function, $z$ is the latent vector of size ($1 \times n_z$), and $y$ is the conditionally generated image which complies with the target conditions $c$ and $s$. Our target can be expressed as training a deep generator network to fit the target mapping function $G\left(z,c,s\right)\rightarrow y$, where the joint distribution $\mathbb{P}(y,c,s)$ is expected to follow the same distribution as $\mathbb{P}(x,c,s)$.

\subsection{Spatially Constrained Generative Adversarial Networks}
In this paper, we propose a generative model called Spatially Constrained Generative Adversarial Networks (SCGAN) to help training a generator network to fit the target mapping function $G\left(z,c,s\right)\rightarrow y$. Our proposed SCGAN consists of three networks shown in Figure~\ref{fig_framework}, which are a generator network $G$, a discriminator network $D$, and a segmentor network $S$. Here we introduce each network individually in detail, define their objective functions, and provide a training algorithm to optimize these networks. 

\label{section_SCGAN}
\noindent\textbf{Generator Network. }
\label{section_generator}
We utilize a generator network $G$ to match our desired mapping function $G\left(z,c,s\right)\rightarrow y$. Our generator takes three inputs which are a latent code $z$, an attribute label $c$, and a target segmentation map $s$. As shown in Figure~\ref{fig_framework}, these inputs are fed into the generator step by step in orders. First, the generator $G$ takes $s$ as input and extracts spatial information contained in $s$ by several downsampling convolutional layers. After that, the convolution result is concatenated with a dimensional expansion of $z$ in channel dimension. After a few upsampling residual blocks (RESBLKUP), $c$ is fed into the generator at last to guide the generator to generate attribute-specific images that contain basic image contents generated from $s$ and $z$. This particular design of $G$ decides the spatial configuration of the synthesized image according to the spatial constraints extracted from $s$. Then $G$ forms the basic structure (\eg background, ambient lighting) of the generated image using the information coded in $z$. After that, $G$ generates the attribute components specified by $c$.


\noindent\textbf{Discriminator Network. }
To obtain realistic results which can hardly be distinguished from the real images, we employ a discriminator network $D$ which forms a GAN framework with $G$. An auxiliary classifier is embedded in $D$ to do a multi-class classification which provides attribute-level and domain-specific information back to $G$. 
$D$ is defined as $D : x \rightarrow \left\lbrace D_d(x),D_c(x) \right\rbrace$, where $D_d(x)$ gives the discrimination results and $D_c(x)$ outputs the probabilities of $x$ belonging to $n_c$ attributes. $D$ and $G$ are two adversarial players in training, which eventually makes $\mathbb{P}(G\left(z,c,s\right),c)$ close to $\mathbb{P}(x,c)$.

\noindent\textbf{Segmentor Network. }
\label{section_methodology_segmentor}
We propose a segmentor network $S$ to provide spatial constraints in conditional image generation. Let $S(\cdot)$ be the mapping function. $S$ takes either real or generated image data as input and outputs the probabilities of pixel-wise semantic segmentation results of size ($H \times W \times n_s$). $S$ can be trained solely using $x$ with its corresponding $s$. When training the other networks SCGAN, the weights in $S$ can be fixed, and $S$ can still provide the gradient information to $G$. Training $S$ separately speeds up the model convergence and reduces the memory usage of the GPUs. 

\subsection{Objective Functions}
\noindent\textbf{Adversarial Loss. }
We adopt a conditional objective from Wasserstein GAN with gradient penalty \cite{gulrajani2017improved}
\begin{equation}
\mathcal{L}_{adv}= {L}_{adv}^{real} + {L}_{adv}^{fake} + {L}_{gp},
\end{equation}
which can be rewritten as
\begin{equation}
\begin{aligned}
\mathcal{L}_{adv}=
& \mathbb{E}_{x}\left[D_{d}\left(x\right)\right]+\mathbb{E}_{z,c,s}\left[D_{d}\left(G\left(z,c,s \right)\right)\right]\\
& + \lambda_{gp}\mathbb{E}_{\hat{x}}\left[\left(\left\|\triangledown_{\hat{x}}D_{d}\left(\hat{x}\right)\right\|_{2}-1\right)^2\right],
\end{aligned}
\end{equation}
where $G\left(z,c,s\right)$ is the generated image conditioned on both attribute label $c$ and segmentation $s$, $\lambda_{gp}$ controls the weight of gradient penalty term, $\hat{x}$ is the uniformly interpolated samples between a real image $x$ and its corresponding fake image $G(z,c,s)$. During the training process, $D$ and $G$ act as two adversarial players that $D$ tries to maximize this loss while $G$ tries to minimize it.

\noindent\textbf{Segmentation Loss }
acts as a spatial constraint to regulate the generator to comply with the spatial information defined by the input semantic segmentation. The proposed real segmentation loss to optimize the segmentor network $S$ can be described as
\begin{equation}
\label{seg_loss_real}
\mathcal{L}_{seg}^{real} = \mathbb{E}_{x,s}[A_s(s, S(x)],
\end{equation}
where $A_s(\cdot, \cdot)$ computes cross-entropy pixel-wisely by
\begin{equation}
A_s(a, b) = -\sum_{i=1}^{H}\sum_{j=1}^{W}\sum_{k=1}^{n_s} a_{i,j,k} \log b_{i,j,k},
\end{equation}
where $a$ is the ground-truth segmentation and $b$ is the estimated segmentation of $a$ of size ($H \times W \times \ n_s$). Taking a real image $x$ as input, estimated segmentation $S\left(x\right)$ is compared with ground-truth segmentation $s$ to optimize the segmentor $S$. When training together with the generator $G$, the segmentation loss term to optimize $G$ is defined as
\begin{equation}
\label{seg_loss_fake}
\mathcal{L}_{seg}^{fake}=\mathbb{E}_{z,c,s}\left[A_s(s, S(G(z, c, s)))\right],
\end{equation}
where the estimated segmentation $S(G(z, c, s))$ is compared with the input segmentation $s$. By minimizing this loss term, the generator is forced to generate fake images which are consistent with the input semantic segmentations $s$.

\noindent\textbf{Classification Loss. }
We embed an auxiliary multi-attribute classifier $D_c$ which shares the weights with $D_d$ in discriminator $D$ except the output layer. The auxiliary classifier $D_c$ takes an image as input and classify the image into independent probabilities of $n_c$ attribute labels. The classification loss for real samples is defined as
\begin{equation}
\label{cls_loss_real}
\mathcal{L}_{cls}^{real}=\mathbb{E}_{x, c} \left[ A_c(c, D_{c}(x) ) \right],
\end{equation}
where ($x,c$) is a pair of real image with its attribute label, $A_c(\cdot, \cdot)$ computes a multi-attribute binary cross-entropy loss by $A_c(a, b) = -\sum_{k}a_k\log(b_k)$ with $a,b$ being two vectors of identical size ($1 \times n_c$). Accordingly, we have the classification loss for fake samples by
\begin{equation}
\label{cls_loss_fake}
\mathcal{L}_{cls}^{fake}=\mathbb{E}_{z,c,s} \left[ A_c(c, D_{c}(G(z, c, s)))\right],
\end{equation}
which takes the fake image $G(z, c, s)$ as input and guides $G$ to generate attribute-specific images according to the classification information learned from real samples.

\noindent\textbf{Overall Objectives }
to optimize $S$, $D$ and $G$ in SCGAN can be represented as
\begin{equation}
\label{full_loss_S}
\mathcal{L}_{S}=\mathcal{L}_{seg}^{real},
\end{equation}
\begin{equation}
\label{full_loss_D}
\mathcal{L}_{D}=-\mathcal{L}_{adv}+\lambda_{cls}\mathcal{L}_{cls}^{real},
\end{equation}
\begin{equation}
\label{full_loss_G}
\mathcal{L}_{G}=\mathcal{L}_{adv}^{fake}+\lambda_{cls}\mathcal{L}_{cls}^{fake}+\lambda_{seg}\mathcal{L}_{seg}^{fake},
\end{equation}
where $\mathcal{L}_{S}$, $\mathcal{L}_{D}$ and $\mathcal{L}_{G}$ are objective functions to optimize $S$, $D$ and $G$.  $\lambda_{seg}$ and $\lambda_{cls}$ are hyper-parameters which control the relative importance of $\mathcal{L}_{seg}$ and $\mathcal{L}_{cls}$ compared to $\mathcal{L}_{adv}$. 

\begin{algorithm}[t]
\caption{Training SCGAN, where $\lambda_{cls}=5$, $\lambda_{seg}=1$, $\lambda_{gp}=10$, $n_{repeat}=5$ and $m=16$.}\label{alg_scgan}
\SetAlgoLined
  Initialize three network parameters $\theta_{G}$ $\theta_{D}$, $\theta_{S}$\;
  \While{$\theta_{G}$ has not converged}{
    \For{$n=1,...,n_{repeat}$}{
      Sample a batch of latent vectors $\{z^i\}_{i=1}^m\sim\mathcal{N}(0,1)$\;
      Sample a batch of $\{x^i,c^i,s^i\}_{i=1}^m$ from $\mathbb{P}_{data}(x,c,s)$\;
      Sample a batch of numbers $\{\epsilon^i\}_{i=1}^m\sim\mathcal{U}(0,1)$\;
      $\{s_t^i\}_{i=1}^m \leftarrow \textbf{shuffle}(\{s^i\}_{i=1}^m)$\;
      \For{$i=1,...,m$}{
        $\tilde{x}^i \leftarrow G(z^i,c^i,s_t^i)$\;
        $\hat{x}^i \leftarrow \epsilon^i x^i+(1-\epsilon^i)\tilde{x}^i$\;
        $\mathcal{L}_{adv}^i\leftarrow D_d(\tilde{x}^i)-D_d(x^i)$\\
        \Indp \Indp ~~~$+\lambda_{gp}(\|\triangledown_{\hat{x}}D_{d}(\hat{x}^i)\|_{2}-1)^2$\;
        \Indm \Indm $\mathcal{L}_{cls}^{real,i}\leftarrow A_c(c^i, D_{c}(x^i))$\;
        $\mathcal{L}_{seg}^{real,i} \leftarrow A_s(s^i, S(x^i))$\;
      }
      Update $D$ by descending its gradient:\\
      \Indp $\triangledown_ {\theta_{D}}
        \frac{1}{m}\sum_i^m{\mathcal{L}_{adv}^{i}+\lambda_{cls}\mathcal{L}_{cls}^{real,i}}$\;
      \Indm Update $S$ by descending its gradient:\\
      \Indp $\triangledown_ {\theta_{S}}
        \frac{1}{m}\sum_i^m{\mathcal{L}_{seg}^{real,i}}$\;
    }
    \For{$i=1,...,m$}{
      $\tilde{x}^i \leftarrow G(z^i,c^i,s_t^i)$\;
      $\mathcal{L}_{adv}^i\leftarrow D_d(\tilde{x}^i)$\;
      $\mathcal{L}_{cls}^{fake,i}\leftarrow A_c(c^i, D_{c}(\tilde{x}^i))$\;
      $\mathcal{L}_{seg}^{fake,i} \leftarrow A_s(s_t^i, S(\tilde{x}^i))$\;
    }
    Update $G$ by descending its gradient:\\
    \Indp $\triangledown_ {\theta_{G}}
    \frac{1}{m}\sum_i^m{(\mathcal{L}_{adv}^i+\lambda_{cls}\mathcal{L}_{cls}^{fake,i}+\lambda_{seg}\mathcal{L}_{seg}^{fake,i})}$\;
    }
    
\end{algorithm}

\begin{figure*}
  \centering
  \includegraphics[width=0.99\textwidth]{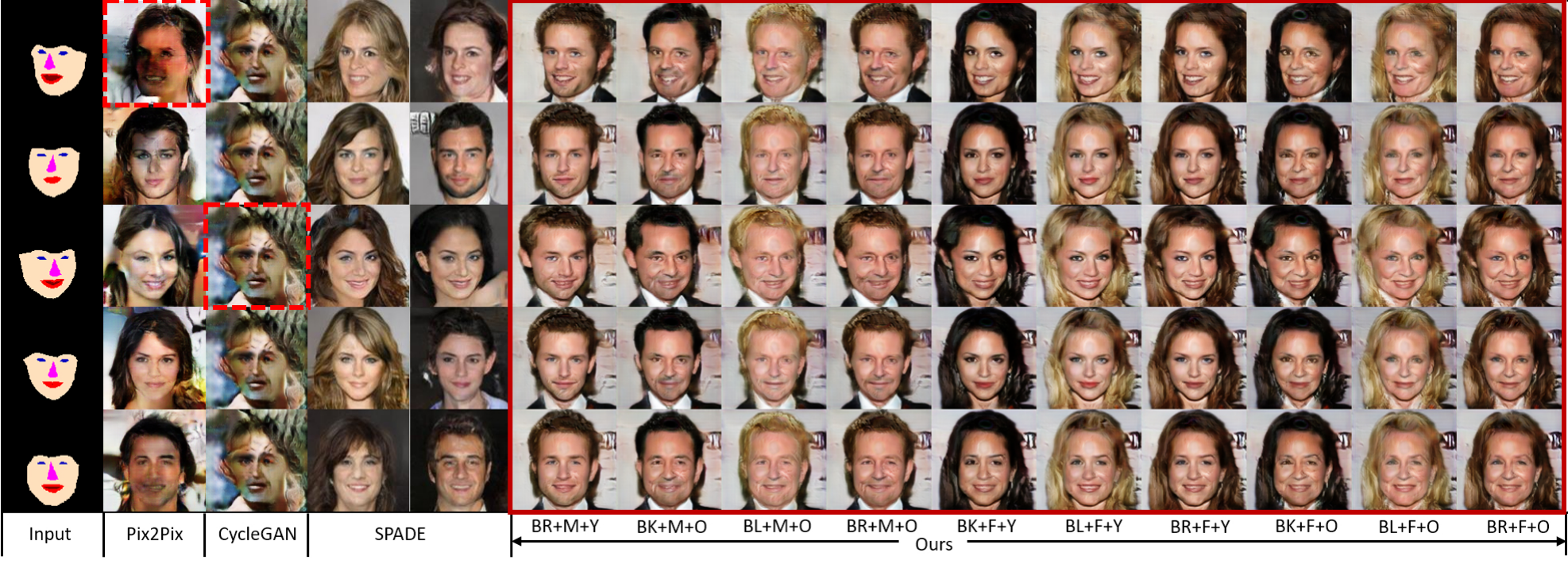}
  \caption{Comparison results on CelebA dataset. Our results are shown in the solid red rectangle. Failure cases of the compared methods are highlighted by the dashed red rectangle. (Abbrev.: BL=Blond Hair, BR=Brown Hair, BK=Black hair, M=Male, F=Female, Y=Young, O=Old.)}\label{fig_compare_celeba}\vspace{-0.3cm}
\end{figure*}

\subsection{Training Algorithm}
Let $\theta_{G}$, $\theta_{D}$ and $\theta_{S}$ be the parameters of networks $G$, $D$ and $S$, respectively. Our objective is to find a converged $\theta_{G}$ with minimized $\mathcal{L}_{G}$. When training the proposed SCGAN, a batch of latent vectors are sampled from a Gaussian distribution $\mathcal{N}(0,1)$ denoted as $\{z^i\}_{i=1}^m$, where $m$ is the batch size. A batch of $x$ with its ground-truth $s$ and $c$ are randomly sampled from the joint distribution $\mathbb{P}_{data}(x,c,s)$ of the target dataset, denoted as $\{x^i, c^i, s^i\}_{i=1}^m$. When selecting target semantic segmentation for $\{x^i\}_{i=1}^m$, $\{s^i\}_{i=1}^m$ are randomly shuffled to obtain a batch of target segmentations $\{s_t^i\}_{i=1}^m$ to be input to $G$. Details can be found in Algorithm \ref{alg_scgan}.

\section{Experiment}
\label{section_experiment}
In this section, we verify the effectiveness of SCGAN on a face dataset and a fashion dataset with both semantic segmentation and attribute label. We show both visual and quantitative results compared with four representative methods, present the spatial interpolation ability of our model in terms of face synthesis.

\subsection{Datasets}

Large-scale CelebFaces Attributes (CelebA) dataset \cite{liu2015faceattributes} contains 202,599 face images of celebrities with 40 binary attribute labels and 5-point facial landmarks. We use the aligned version of face images and select 5 attributes including black hair, blond hair, brown hair, gender, and age in our experiment. This dataset doesn't provide any ground-truth semantic segmentation for the face images. To obtain the semantic segmentation, we apply Dlib \cite{dlib09} landmarks detector to extract 68-point facial landmarks from the faces images, which separate facial attributes into six different regions. By filling those regions with corresponding semantic index pixel-wisely, semantic segmentations are created. 

Large-scale Fashion (DeepFashion) dataset \cite{liu2016deepfashion,iccv2017fashiongan} is a large-scale clothing database, which contains over 800,000 diverse fashion images ranging from well-posed shop images to unconstrained photos from customers. In our experiment, we use one of the subsets particularly designed for the fashion synthesis task, which selects 78,979 clothing images from the In-shop Clothes Benchmark associated with their attribute labels, captions, and semantic segmentations. We use the 18-class color attributes and the provided semantic segmentation in our experiment.

\subsection{Compared Methods}
Pix2Pix \cite{isola2016image} and CycleGAN \cite{zhu2017unpaired} are two popular image-to-image translation me\-thod, which can take semantic segmentation as input and synthesize realistic images. Pix2Pix requires paired images while CycleGAN is trained in an unpaired way. We also compare our method with a most recent state-of-the-art method named SPADE \cite{park2019semantic} which can generate images given semantic segmentations following the styles/modalities of input images. In our experiment, we use the official implementation released by the authors, train their model on our target datasets, and try our best to tune the parameters to deliver good results.

\subsection{Spatially Constrained Face Synthesis}
We first provide comparison results on CelebA in Figure \ref{fig_compare_celeba}. The input segmentations are shown in the leftmost column, and the results of Pix2Pix and CycleGAN are shown in the next two columns. The visual quality generated by Pix2Pix is low, and CycleGAN suffers a mode collapse issue that their model only gives a single output no matter the input segmentation. One possible reason is that translating facial segmentation to realistic faces is essentially a one-to-many translation, however, those two image-to-image translation methods both assume a one-to-one mapping between input and target domains. Especially for CycleGAN, their cycle-consistency loss which seeks to maintain the contents during a cycle translating forward and backward tends to enforce the one-to-one mapping. When a face image is translated into its semantic segmentation, it is barely possible to translate it back to the original face due to the information lost in the many-to-one translation. The results of SPADE are presented in Column 4 and 5 in Figure \ref{fig_compare_celeba}. SPADE can generate diverse faces given fixed segmentation as inputs, but the attributes of the generated faces are randomized despite providing ``style images'' to the encoder. Since face images are similar to each other in structures, it violates the style-based assumption of SPADE. Our proposed SCGAN could always produce reliable and high-quality results. It is worth noting that inputting randomly sampled latent vectors can result in diverse images with different backgrounds and details in segmentation-to-image synthesis. Due to the high-frequency signal from boundaries of attributes in semantic segmentation, our SCGAN could produce a large number of sharp details which makes the results more realistic compared to all the other methods. In summary, SC\-GAN enjoys superiority in terms of diverse variations, controllability, and realistic high-quality results over the other methods.

\begin{figure}[t]
  \centering
  \includegraphics[width=0.48\textwidth]{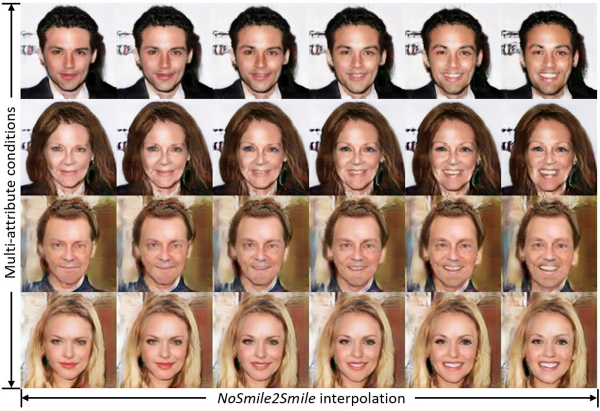}
  \caption{\emph{NoSmile2Smile} facial expression interpolations. Each row shows a group of interpolated results between a not smiling face and a smiling face with a specific attribute label and a fixed latent vector. }\label{fig_interpolation_nosmile2smile}\vspace{-0.2cm}
\end{figure}

\subsection{Interpolation Abilities}
\label{sec_interpolation}
\begin{figure}[t]
  \centering
  \includegraphics[width=0.48\textwidth]{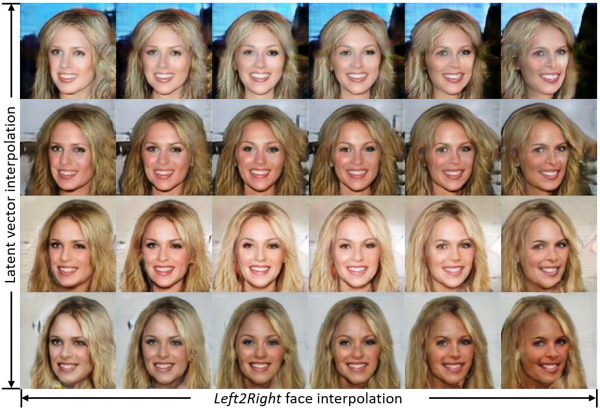}
  \caption{Two-dimension interpolation results in latent space and between \emph{Left2Right} faces. Each column presents the results of interpolated latent vectors, and each row shows the interpolation results on facial orientations. }\label{fig_interpolation_left2right}\vspace{-0.2cm}
\end{figure}

Beyond face synthesis, our proposed SCGAN can control the face orientation and facial expressions of the synthesized faces by feeding corresponding semantic segmentations as guidance. To synthesize faces of every intermediate state between two facial orientations and expressions, corresponding semantic segmentations of those intermediate states are needed. It is difficult to obtain such intermediate segmentations that numeric interpolation between two segmentations only results in a fade-in and fade-out effect. Instead, we interpolate every intermediate state on $x$-$y$ coordinates of facial landmarks instead of segmentation domain. We then construct semantic segmentations from those landmarks to obtain spatial-varying semantic segmentation. As shown in Figure \ref{fig_interpolation_nosmile2smile} and \ref{fig_interpolation_left2right}, SCGAN generates intermediate faces from not smiling face to smiling face (\emph{NoSmile2Smile})and from left-side face to right-side face (\emph{Left2Right}). Interpolations on latent vectors are also shown in Figure \ref{fig_interpolation_left2right}.

\begin{figure*}[t]
  \centering
  \includegraphics[width=0.99\textwidth]{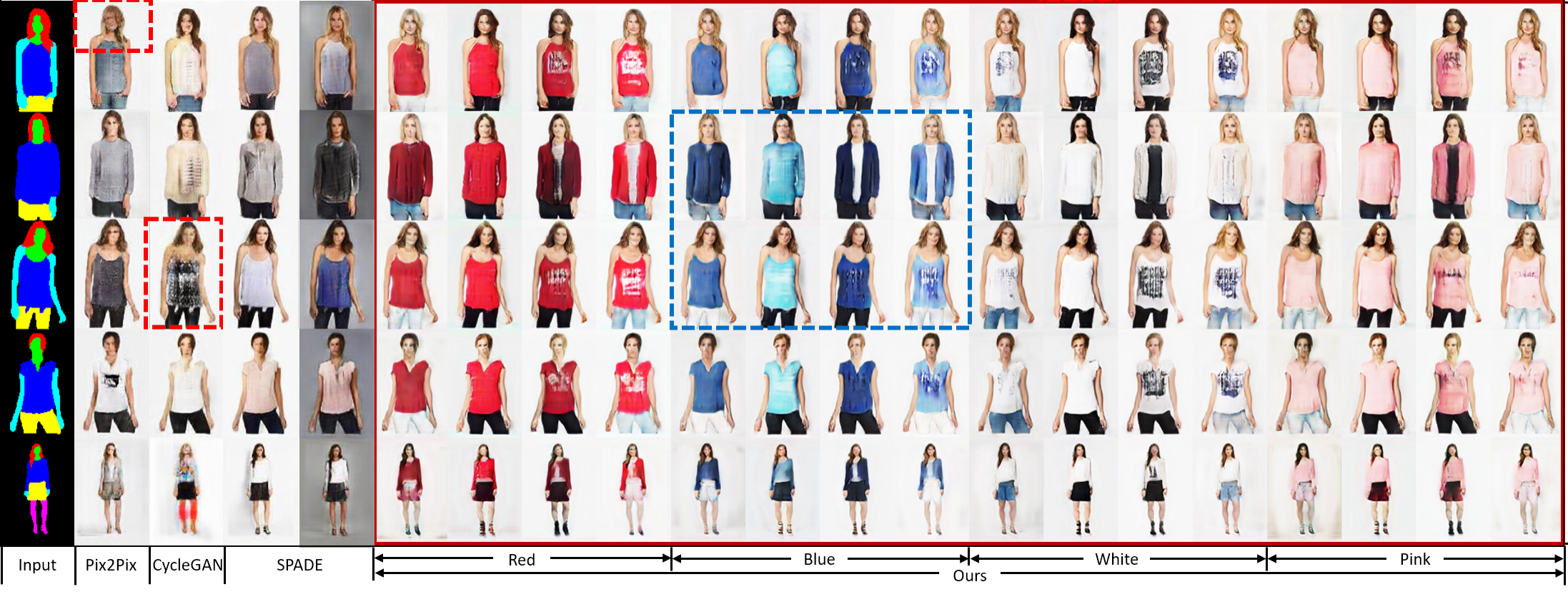}
  \caption{Comparison with Pix2Pix, CycleGAN and SPADE on DeepFashion dataset. Their failure cases are highlighted in the dashed red rectangle, while the dashed blue rectangle highlights the representative diverse results generated by our proposed SCGAN.}\label{fig_compare_fashion}\vspace{-0.2cm}
\end{figure*}

\begin{table}[t]
\caption{Quantitative evaluation on CelebA and DeepFashion dataset using Fr\'{e}chet Inception Distance (FID), mean IoU (mIoU) and pixel accuracy (pAcc). N/A indicates mode collapse}\label{tab_accuracy}
\centering
\begin{tabular}{|p{1.8cm}|P{0.6cm}|P{0.6cm}|P{0.6cm}|P{0.6cm}|P{0.6cm}|P{0.6cm}|}
  \hline
  ~ & \multicolumn{3}{c|}{CelebA} & \multicolumn{3}{c|}{DeepFashion} \\
  \hline
  Methods &  FID & mIoU & pAcc &  FID & mIoU & pAcc \\
  \hline
  CycleGAN \cite{zhu2017unpaired} & N/A & N/A & N/A & 30.1 & 63.26 &  82.21 \\
  Pix2Pix \cite{isola2016image}  & 20.4 & 78.71 &  98.05 & 24.4 & 65.41 &  82.91\\
  SPADE \cite{park2019semantic} & 18.5 & 74.76 & 97.82 & 20.2 & 75.80 & 83.10\\
  \hline
  \textbf{SCGAN} & \textbf{10.2}  & \textbf{79.11} & \textbf{98.95} & \textbf{19.8} & \textbf{77.20} & \textbf{83.23}\\
  \hline
\end{tabular}\vspace{-0.2cm}
\end{table}

\subsection{Spatially Constrained Fashion Synthesis}
Comparison results on the DeepFashion dataset presented in Figure \ref{fig_compare_fashion} also demonstrate the advantages of our proposed SCGAN over the other methods. Similar to Figure \ref{fig_compare_celeba}, the input segmentation, results of Pix2Pix, CycleGAN and SPADE are shown in the left five columns. The images in the large solid red rectangle are our results from SCGAN with both semantic segmentation and attribute labels and latent vector as inputs. Different from the results on CelebA dataset, image-to-image translation methods are capable of producing acceptable results on the DeepFashion dataset, because the intrinsic one-to-many property in the DeepFashion dataset is not as strong as in the CelebA dataset. In the DeepFashion dataset, the ability of shape preserving becomes more important than general visual discrimination. Their results also lack attribute-level controllability and variations on fashion detail as our results highlighted by the dashed blue rectangle. With our semantic segmentation as the spatial constraints, SCGAN can generate fashion images controlled by the input color labels and semantic segmentation, while the input latent vectors encode variant fashion style (\eg cardigans, T-shirts), diverse pants and shoes, and different color shades and saturation (\eg dark blue, light blue).



\begin{figure}[t!]
  \centering
  \includegraphics[width=0.46\textwidth]{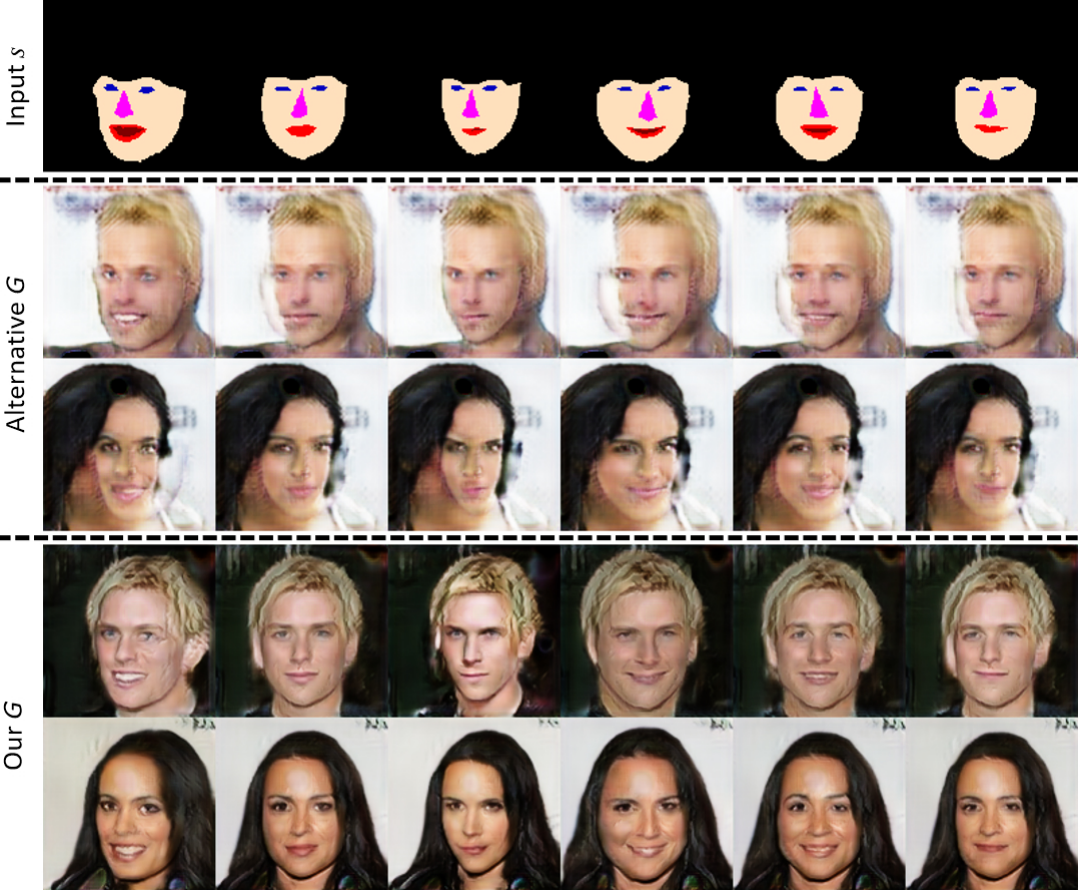}
  \caption{An ablation study on generator configurations. Our proposed generator structure solves the foreground-background mismatch problem suffered by the alternative generator which inputs all the conditions at once.}\label{fig_ablation_g}\vspace{-0.3cm}
\end{figure}

\subsection{Quantitative Evaluation}
To quantitatively evaluate the effectiveness of spatially constrained image generation, we use Fr\'{e}chet Inception Distance (FID) \cite{heusel2017gans} to evaluate the fidelity of the generated images. FID measures the distance between real and synthesized data in their Inception embeddings. We also adopt metrics of mean IoU (intersection over union) and pixel accuracy to examine the spatial consistency between the input semantic segmentation and the generated images from the generator, which are commonly used when evaluating segmentation algorithms. We run the experiment for five times and report the averaged results compared with the image-to-image translation methods of CycleGAN, Pix2Pix and SPADE. As shown in Table \ref{tab_accuracy}, our SCGAN achieves the best performance on both CelebA and DeepFashion datasets. Our method is capable of generating realistic images with diversity as well as make those images comply with the input semantic segmentations accurately.

\begin{figure}[t]
  \centering
  \includegraphics[width=0.48\textwidth]{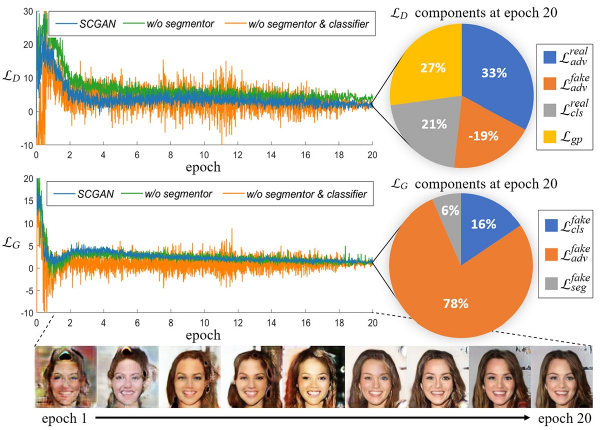}
  \caption{An ablation study on model convergence. Losses during training are plotted together and the intermediate generated samples are shown.}\label{fig_convergence}\vspace{-0.3cm}
\end{figure}

\subsection{Ablation Study on Generator Configuration}
As described in Section \ref{section_generator}, the generator of our proposed SCGAN takes three inputs, a semantic segmentation, a latent vector, and an attribute label step by step in order that the contents in the synthesized image should be decoupled well to be controlled by those inputs. Otherwise, those inputs may conflict with each other and fail to generate the desired results. To demonstrate that, we conduct an ablation study to compare with an alternative generator that takes all the inputs and concatenates them together at the same time. We refer to this variant of generator network as the alternative $G$. As shown in Figure \ref{fig_ablation_g}, severe foreground-background mismatches happen in the results of alternative $G$ that the facial components regulated by the input segmentation cannot be merged correctly with the skin color or hairstyle determined by the latent vector. Our particularly designed generator could successfully decouple the contents of synthesized images into controllable inputs and generate variations on other unregulated contents. 

\subsection{Ablation Study on Model Convergence}
Our proposed SCGAN converges fast and stably due to the introduction of the segmentor and the auxiliary classifier. We conduct an ablation study on model convergence by removing segmentor and auxiliary classifier. Figure \ref{fig_convergence} shows the losses of generator and discriminator during the training process on \emph{CelebA} dataset. The blue plots are the losses of the proposed SCGAN. Green plots are the losses after removing the segmentor network. The orange plots show the losses after removing both the segmentor network and the embedded auxiliary classifier, while all the other things such as model architecture and hyper-parameters are kept unchanged. As observed from this figure, the training process of our SCGAN is much more stable with less vibration on losses. The convergence of SCGAN happens faster and its final loss is smaller than the other two ablation experiments. The bottom part in Figure \ref{fig_convergence} shows the intermediate generated samples that improve gradually as the model converges.

\section{Conclusions}
In this paper, we proposed SCGAN to introduce spatial constraints to face and fashion image synthesis. Extensive experiments compared with other popular generative models on CelebA face dataset and DeepFashion datasets demonstrated that the proposed SCGAN was capable of controlling spatial contents, specifying attributes, and generating diversified images. We particularly designed the generator to take semantic segmentations, latent vectors, and attribute labels step by step to solve the foreground-background mismatch problem. In summary, our method is a simple yet effective variant of the GAN, which could be easily adapted to recent high-resolution GAN-based image generation models.